%% file: ms.tex
\documentclass[letterpaper, 10 pt, conference]{ieeeconf}  

\IEEEoverridecommandlockouts                              

\usepackage{newtxtext,newtxmath}
\usepackage{algorithm}
\usepackage[noend]{algpseudocode}
\usepackage{tabularx,ragged2e,booktabs}
\usepackage{tabulary}
\usepackage{bm}
\usepackage{cite}
\usepackage{soul}
\usepackage{float}

\usepackage{hyperref}

\usepackage{graphics} 
\usepackage{epsfig} 
\usepackage{mathtools}
\usepackage{times} 
\usepackage{xcolor}
\usepackage{multicol}
\usepackage{multirow}
\usepackage{colortbl}
\usepackage{lipsum}

\title{\LARGE \bf
Open-VICO: An Open-Source Gazebo Toolkit for Vision-Based Skeleton Tracking in Human-Robot Collaboration
}
\author{Luca Fortini$^{1,2}$, Mattia Leonori$^{1}$, Juan M. Gandarias$^{1}$, Elena De Momi$^{2}$ and Arash Ajoudani$^{1}$ 
	\thanks{$^1$Human-Robot Interfaces and physical Interaction Laboratory,  Istituto Italiano di Tecnologia, Genoa, Italy, Email: luca.fortini@iit.it}%
	\thanks{$^2$Department of Electronics, Information and Bioengineering, Politecnico di Milano, Milano, Italy.}%
\thanks{Github~Repo:~\url{https://gitlab.iit.it/hrii-public/open-vico}}%
}%

\begin{document}
\maketitle
\thispagestyle{empty}
\pagestyle{empty}


\input{ms1}
\label{Sec::abstact}

\section{Introduction}
\input{ms2}
\label{Sec::introduction}

\section{Toolkit Overview}
\input{ms3}
\label{Sec::overview}

\section{Application Examples}

\input{ms4}

\label{sec:examples}

\section{Conclusions}
\input{ms5}
\label{sec:conclusions}

\section{Acknowledgements}
\label{sec:acknowledgements}
\input{ms6}

\bibliographystyle{IEEEtran.bst}
\bibliography{ms.bib}








\end{document}

%% file: ms1.tex
\begin{abstract}
Simulation tools are essential for robotics research, especially for those domains in which safety is crucial, such as Human-Robot Collaboration (HRC). However, it is challenging to simulate human behaviors, and existing robotics simulators do not integrate functional human models. This work presents Open-VICO, an open-source toolkit to integrate virtual human models in Gazebo focusing on vision-based human tracking. In particular, Open-VICO allows to combine in the same simulation environment realistic human kinematic models, multi-camera vision setups, and human-tracking techniques along with numerous robot and sensor models thanks to Gazebo. The possibility to incorporate pre-recorded human skeleton motion with Motion Capture systems broadens the landscape of human performance behavioral analysis within Human-Robot Interaction (HRI) settings.
To describe the functionalities and stress the potential of the toolkit four specific examples, chosen among relevant literature challenges in the field, are developed using our simulation utils: i) 3D multi-RGB-D camera calibration in simulation, ii) creation of a synthetic human skeleton tracking dataset based on OpenPose, iii) multi-camera scenario for human skeleton tracking in simulation, and iv) a human-robot interaction example. 
The key of this work is to create a straightforward pipeline which we hope will motivate research on new vision-based algorithms and methodologies for lightweight human-tracking and flexible human-robot applications.
\end{abstract}

%% file: ms2.tex
The future of robotics envisions a world where the close coexistence of robots and humans is a reality~\cite{Feil2011, Ajoudani2018, Matheson2019}. 
In this scenario, safety must be guaranteed, and future robotic systems must be provided with advanced tools to enable high-accuracy human tracking and ensure human safety. 

Human modeling and tracking is a fundamental research topic in multiple industries such as sports~\cite{Pers2001, Barris2008}, healthcare~\cite{Zhou2008, Achilles2016, Ruiz2021}, or entertainment~\cite{Bregler2007, Shere2021}; and is undergoing an enormous momentum in robotics due to the current trends of the discipline and socio-economical demands.  
Existing technologies for human tracking, also called motion capture (MoCap) systems, rely on different information sources such as acceleration data provided by e.g., Inertial Measurement Units (IMUs)~\cite{Roetenberg2009}, beacons~\cite{Arora2005}, or Ultra-Wide Band (UWB) signals~\cite{Chang2010}. Nonetheless, visual perception systems are the most common approach in this regard. These systems can be classified as marker-based~\cite{Nagymate1970} or marker-less~\cite{Cao2021, Chen2017}).  
The performance of marker-less human tracking systems is relatively poor, and the most utilized human tracking techniques regard marker-based vision systems. However, these systems are limited to particular setups due to outrageous expenses or the necessity of wearing specialized suits. 

\begin{figure}
    \centering
    \includegraphics[width=0.85\columnwidth]{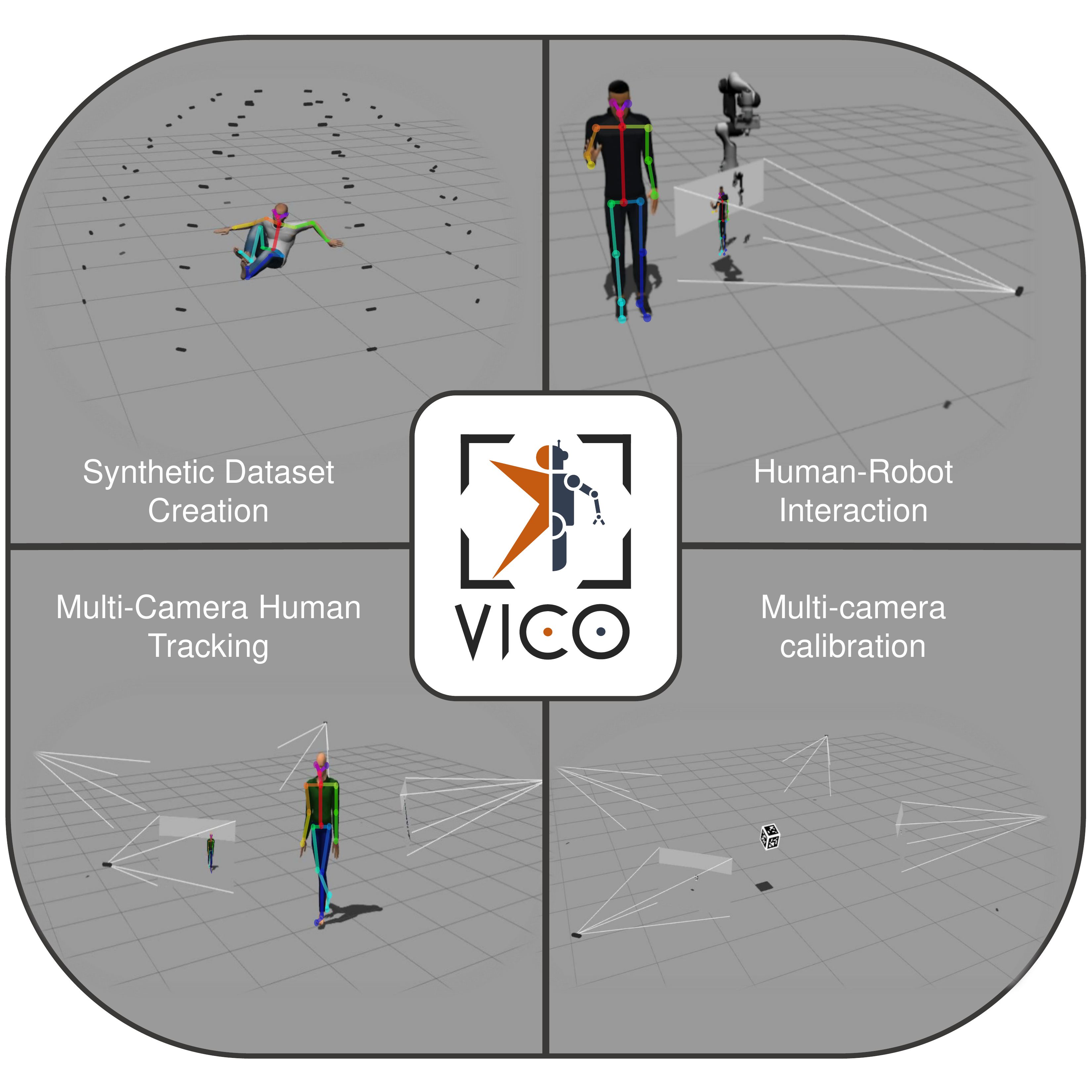}
    \caption{Illustration of Open-VICO possible usages, an open-source Gazebo toolkit for the integration of 3D human models and multi-vision systems within a robotic simulator environment.}
    \label{fig:digest}
\end{figure}

On the other hand, simulation tools provide an excellent instrument to develop and test methodologies and algorithms without compromising hardware integration and safety. In robotics, simulation is essential for numerous and well-known reasons. In~\cite{Choi2021} a detailed roadmap with specific requirements and suggestion for developing simulation environments in robotics is depicted. To specifically tackle the points raised on human-in-the-loop simulation, this work presents Open-VICO (see Fig.~\ref{fig:digest}), an Open-Source Gazebo toolkit conceived for HRI.


\begin{figure*}
    \centering
    \includegraphics[width=0.85\textwidth]{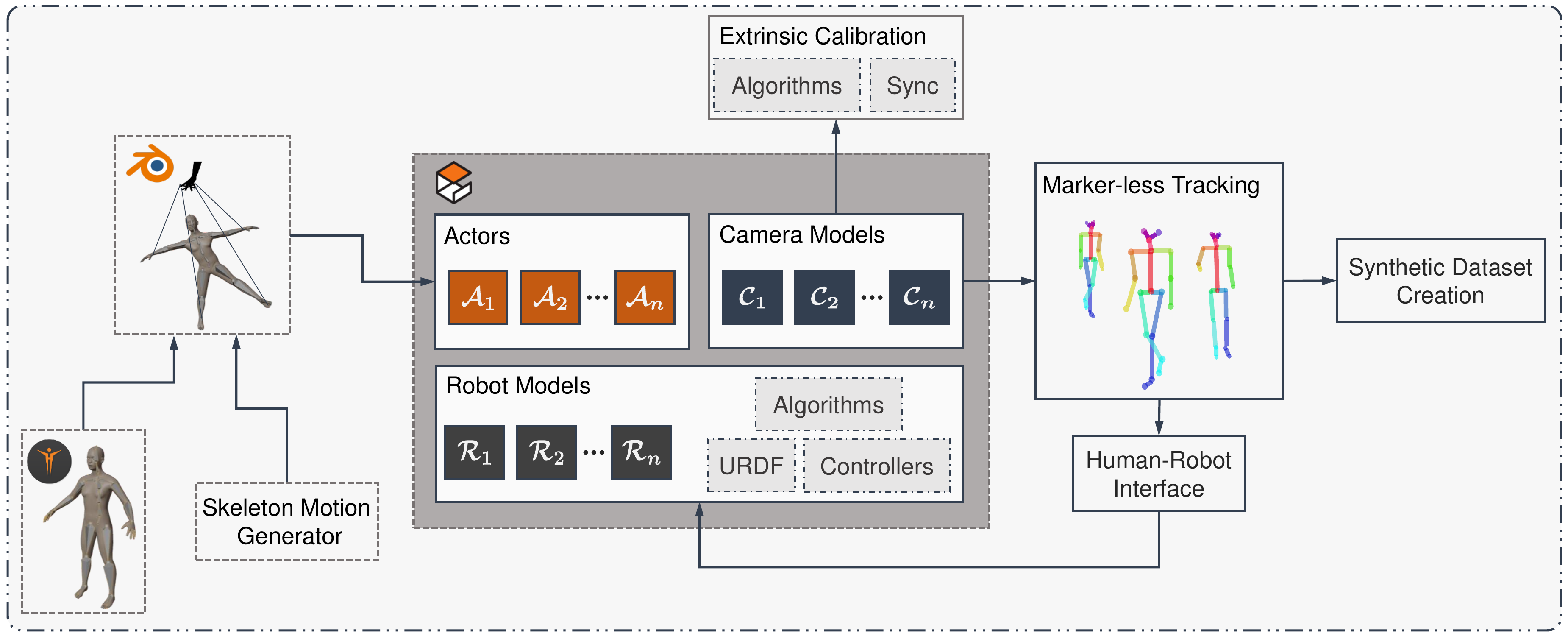}
    \caption{Software and Data structure conceptualization of Open-VICO. The toolkit provides the utils to integrate animated actors in Gazebo thanks to \textit{Blender}, the Makehuman software and a skeleton motion generator (e.g., a MoCap system). Inside Gazebo, Open-VICO ensures a simulated interaction between actors, cameras and robotic models.}
    \label{Fig:structure}
\end{figure*}

Gazebo~\footnote{\url{http://gazebosim.org/}} is one of the most popular 3D ROS-based simulators for robotic environments and systems, complete with dynamic and kinematic physics and a pluggable physics engine. As a practical and manufacturer-independent software, Gazebo offers a rich environment for the rapid development and testing of complex robot systems. Lately some features for human simulation have been implemented however the process is still cumbersome and poorly customisable. Open-VICO tackles this challenge by integrating virtual human kinematic models within the Gazebo framework in a smoother way enhancing the simulation possibilities for the user. 

Moreover, the toolkit allows the spawn of multiple camera models along with vision-based human skeleton tracking methods. In particular, the contributions of this paper are the following: 

\begin{itemize}
	\item Presentation of Open-VICO as an open-source Gazebo toolkit for empowering human kinematic models inside a robotic simulator.
	\item Description of the toolkit software architecture and modeling pipeline.
	\item Illustration of the following four examples to demonstrate the potential of the tool:
	\begin{enumerate}
		\item 3D camera calibration in a simulated multi-camera environment.
		\item Creation of synthetic dataset of human motions in simulation using a simulated RGB-D camera model and OpenPose.
		\item Creation of a multi-RGB-D camera setup with OpenPose that allows the evaluation of the system and enables the development of multi-sensor fusion algorithms in simulation.
		\item An HRI environment in which pre-recorded human motions with a MoCap are integrated within Open-VICO to create virtual human motions to teleoperate a virtual robotic arm.
	\end{enumerate}
\end{itemize}

The rest of the paper is structured as follows: Section~\ref{Sec::overview} describes the general features and relevance of the toolkit, and the software architecture and dependencies. Section~\ref{sec:examples} presents four showcases to demonstrate the features of the tool for different applications considering relevant research challenges in the field. Finally, the conclusions are included in section~\ref{sec:conclusions}.

%% file: ms3.tex

This section describes the main relevance, characteristics and software architecture of Open-VICO. Fig.~\ref{Fig:structure} illustrates a general overview of the data and software architecture. 
Open-VICO comprises a series of tools working as a ROS-based interface and aiming at increasing the humans' presence in the Gazebo platform that so far has been used almost exclusively by the robotics community.
Our code and further documentation is available at \url{https://gitlab.iit.it/hrii-public/open-vico}.

\subsection{Relevance and Potential Applications}
\label{subsec:potential_app}
We list here a series of literature research areas which we believe will benefit from using Open-VICO.

\subsubsection{Extrinsic Calibration of Multi-Sensor Vision Systems}
\label{subsubsec:calibration_app}
In computer vision applications, using a single camera considerably limits the working area, especially when dealing with RGB-D cameras with a narrow Field Of View (FoV). Moreover, other factors affect the performance of single-camera settings, such as occlusions or possible limited robustness. Hence, using multiple cameras to deal with these issues is very appealing. 

Such systems need to be calibrated in order to have the data perceived from the multiple cameras represented w.r.t. the same reference frame. This process is called extrinsic calibration and estimates the relative poses between the cameras. The most popular solutions consist of using a calibration pattern to be shown simultaneously to the cameras trying to optimize the reprojection error \cite{Zhang2000, Sarmadi2019}. Nevertheless, the number of possible solutions is considerable, and choosing the best fit for an application may not be trivial. In addition, there are some extra challenges, such as combining different types of cameras.

Especially when developing and testing a new algorithm or comparing it to others, the opportunity to assess the result's accuracy could be of great help. Open-VICO offers a simple pipeline to spawn multiple cameras in predefined scriptable configurations and patterns, randomly move a calibration prop and synchronize the images of the different cameras using the rosbag tools. In particular, the possibility to have ground-truth data resulting in a readable and standard format is of interest for the computer vision community. In addition, extrinsic calibration algorithms can be implemented and tested in simulation without expending hardware resources and wasting time locating markers in several positions or moving them in the scene.

\subsubsection{Synthetic Human Motion Dataset Creation}
\label{subsubsec:dataset_app}
Many applications hold fundamental challenges for estimating human pose, shape, and motion from images and videos. Recent advances in 2D human pose estimation use large amounts of manually-labeled training data for learning convolutional neural networks (CNNs). Labeling all these data is time-consuming and challenging to extend. Moreover, manual labeling of the 3D pose, depth, and motion attributes is impractical. It has been proved that the use of synthetically-generated datasets guarantees good performances in training networks opening new possibilities for the use of cheap, potentially limitless datasets ~\cite{Varol2017, Varol2021}. 

Open-VICO is integrated within common robotic ROS-based toolsets, providing familiar means by which roboticists can build labeled RGB and RGB-D synthetic datasets for their work. This tool is designed for minimal user involvement and maximum flexibility during the data generation process. The user can specify arbitrary motion plans for various objects in the scene, camera position, simulated frame rate, actor aspect and dress-code, background, and luminosity level, among other attributes. Another fundamental potential of Open-VICO is to improve the performance of action recognition by creating synthetic data in cases where the actual data is limited, e.g., domain mismatch between training/tests such as viewpoints or low-data regime.

\subsubsection{Human Tracking with Multi-Sensor Vision Systems}
\label{subsubsec:multi_tracking}
There is a need within human movement sciences for a markerless mocap system, which is easy to use and sufficiently accurate to evaluate motor performance. In the last decade, deep-learning-based markerless motion capture systems attracted much research interest, obtaining sufficiently fair results in 2D human tracking. Nevertheless, we are still far from having a stable, accurate, and hardware-compatible 3D skeleton tracking system that competes with flagship marker-based systems. Among these, OpenPose~\cite{Cao2021} is one of the most popular open-source pose estimation approaches. A few attempts were made in this sense to extend OpenPose, lifting it to the third dimension. The most immediate and trivial solution would be to use RGB-D cameras. If the 2D joint location is known, a simple lockup in the depth cloud is acceptable. 

However, depth retrieval still has several issues: low resolution, noise, and occlusions can introduce jerks and jumps in the output, especially when it comes to extremities or at great distances. This issue becomes even more problematic if the 2D joint estimation is not perfect in the first place. As a consequence joining the information of multiple devices using triangulation techniques~\cite{Nakano2020} or fusion algorithms~\cite{Nguyen2021} seems to be a suitable solution for the aforementioned problems.
In this regard, Open-VICO offers the perfect environment for testing and developing these solutions with a fast and direct comparison with ground truth data.

\subsubsection{Human-Robot Interaction in Simulation}
In industrial environments, the collaboration between humans and robots is considered a profitable and almost required strategy to increase productivity and decrease the cost of production. This strategy benefits from combining both the robot's fast repetition and high production capabilities and the operator's reasoning, reacting, and planning abilities. In such a scenario, aiming at guaranteeing an efficient response of the robotic partner requires the constant monitoring of the human actor position and intention. Reliability and accuracy have always favored wearable systems in these kinds of contexts. However, the discomfort caused by the protracted wearing was determinant in arousing a mild enthusiasm in the industrial world. Vision-based solutions have the advantage of immediacy, cost, and freedom of movement.

On the other hand, they lack robustness and suffer occlusions. To overcome these inconveniences is essential to optimize the cameras' framing and the agents' relative position. Moreover, filtering algorithms and techniques may be used to obtain a more desirable behavior.

The research studies in~\cite{Rolley2018, Martin2019} faced issues and took compromises in teleoperating a robot relying on an OpenPose detection system. To overcome jerks and jumps of the robotic counterpart, the authors chose heuristic solutions such as euclidean filters on subjects' links or predetermined operating area of the agent to match the camera Field of View (FoV).

In this regard, the Gazebo ROS-integrated simulation environment offers plenty of options for testing and prototyping these scenarios. Although Gazebo is a physics engine that allows simulating dynamic behaviors, this option has not been yet exploited in the presented toolbox due to its considerable complexities but is the next step of continuous software integration and development within Open-VICO.

\subsection{Human 3D Model Definition and Integration}
\label{subsec:human_model}
This section explains how to build an animated human body model for Gazebo. First, a 3D human model is created and rigged using MakeHuman~\footnote{\url{http://www.makehumancommunity.org/}}, an open-source digital human modeling (DHM) software that offers high detail features to personalize the avatar. Rigging refers to creating the bone structure of a 3D model. The model can then be easily imported in the 3D computer graphics software Blender \footnote{\url{https://www.blender.org/}} using an embedded plugin (see Fig.~\ref{Fig:structure}. This step is still necessary to retarget the 3D model, namely bringing it to life repurpusing previously acquired mocap data as a marionette in .fbx or .bvh extension. Shortly, it will be possible to retarget a Collada model directly in Gazebo, although this feature is still under development. After retargeting, a Collada file should be exported from Blender and spawned in the Gazebo simulation environment through the ``actor'' class.

To allow a cross-comparison of deep-learning-based MoCap systems as anticipated in section~\ref{subsubsec:multi_tracking} and allow a smooth and natural rigging procedure, the skeleton model should meet a certain number of requirements. It should be complex enough in terms of degrees of freedom to assure a natural movement of the avatar and, at the same time, have the fundamental keypoints to guarantee an harmonic joint comparison with most of the 2D deep learning-based MoCap systems, which mainly rely on MPI, COCO, and BODY 25 models. Although rough in the chest's links estimation, the latter offers the chance to track the hands; for this reason, the hands were also added to the rig. If not retargeted, they will follow the wrist parent joint trajectory.

\subsection{Camera Model Integration}
\label{subsec:camera_model}
Especially in cases relying on depth sensors to lift to 3D markerless 2D human tracking algorithms, the sensor performance affects the estimation outcome significantly. Moreover, different camera brands have different FoV or ROS topics names in their ROS wrapper if they have one. Open-VICO provides guidelines to integrate different camera models, allowing it to operate at a higher level and easily merge information from different sensors. So far, three camera models have been integrated within the framework, the Kinect v2, the Realsense D435, and the ZED2. Nevertheless, more models will be integrated as the toolkit is being developed.

\subsection{Skeleton Tracking Method}
\label{subsec:skeleton_tracking}
The apparent advantages that markerless solutions offer in human tracking generated high interest in the research community. As a result, many methodologies and systems are continuously under development, claiming to be the best so far. Open-VICO is the ideal environment to cross-compare and fuse the results of these systems offering the footprint of a bottleneck custom ROS message for joint position and name harmonization compatible with the human landmarks described in sectio~\ref{subsec:human_model}. In~\cite{Rapczynski2021} there are some guidelines on how to perform this harmonization procedure among different training datasets (e.g., COCO) typically used in deep learning-based skeleton tracking systems. Open-VICO's structure allows the user to easily append additional tracking algorithms to the default list, enriching the possible comparison combinations.


\begin{figure}
    \centering
    \includegraphics[width=0.75\columnwidth]{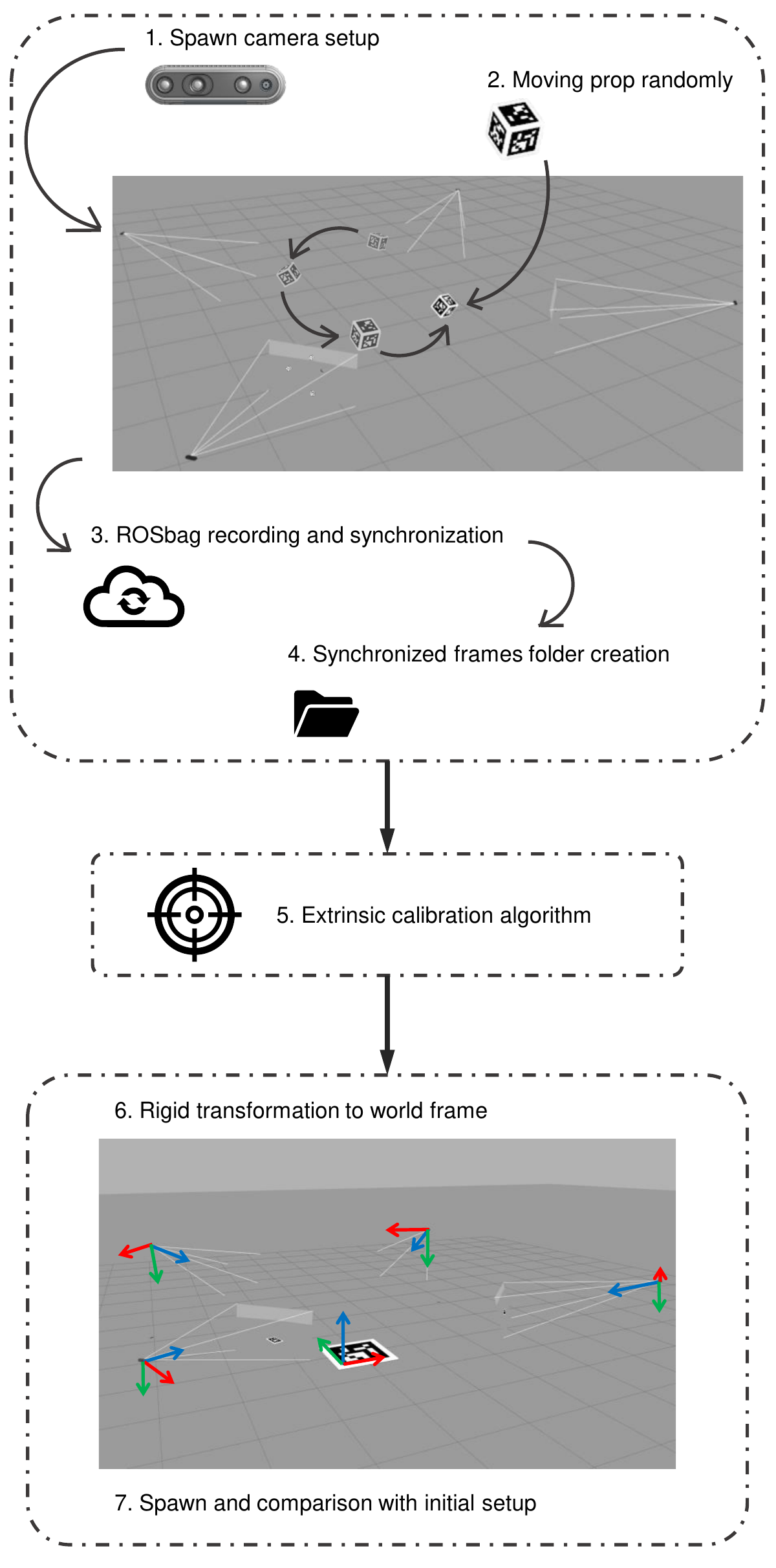}
    \caption{The multi-camera calibration routine in Open-VICO is defined by the seven steps presented in the figure: 1. The multi-camera setup is spawned in the Gazebo world. 2. The virtual prop is spawned in different random positions inside a predefined workspace. 3. The data obtained by the cameras is recorded with a rosbag and synchronized. 4. Folders with synchronized frames are created automatically. 5. The data feeds an extrinsic calibration algorithm (e.g.,~\cite{Sarmadi2019}). 6. Calculation of the rigid transformations from each camera w.r.t. the world frame. 7. Spawn the calibrated setup and comparison with the ground truth to evaluate the performance of the calibration algorithm.}
    \label{Fig:calibration}
	\vspace{-0.5cm}
\end{figure}

%% file: ms4.tex
This section presents four use-cases using Open-VICO utils to demonstrate the proposed framework's potential. Note that the methods presented in this section are not novel per se as they are not intended to be the contribution and aim of this paper. On the contrary, this section implements existing solutions that benefit from the tools provided by Open-VICO to take advantages of working in simulation.

\subsection{Multi-Camera Calibration}
\label{subsec:multi-camera}

An application example of Open-VICO regarding its use for multi-camera settings extrinsic calibration evaluation is presented in this section.
To show the features and the potential of the Open-VICO tool, we employ the extrinsic calibration algorithm described in~\cite{Sarmadi2019} using a synthetic environment and based on Aruco markers detection. That work presented a quantitative evaluation of the method based on tracking custom-shaped reflective markers attached to the cameras to track their pose in the space using an optoelectronic MoCap system. This solution, however, might be considered naive and prone to errors, while working in simulation allows directly comparing the results with the initial configuration.

We picture a scenario where two calibration props are to be compared to test the best fit for the calibration procedure. The steps of the calibration-and-evaluation routine are outlined in Fig.~\ref{Fig:calibration}.
The calibration routine defines the seven steps specified in the figure.

Suppose the coordinate frame of reference is not attached to the camera frame but rather a world coordinate frame attached to some object. In that case, a rigid body transformation (rotation and translation) relates the camera coordinate frame to the world coordinate frame. To obtain this, we used an additional ArUco marker spawned in the gazebo world to overlap the initial configuration setup with the calibrated one (see Fig.~\ref{Fig:calibration}).

\begin{figure}
    \centering
    \includegraphics[trim=7.0cm 3.0cm 7.0cm 3.5cm, clip, width=0.9\columnwidth]{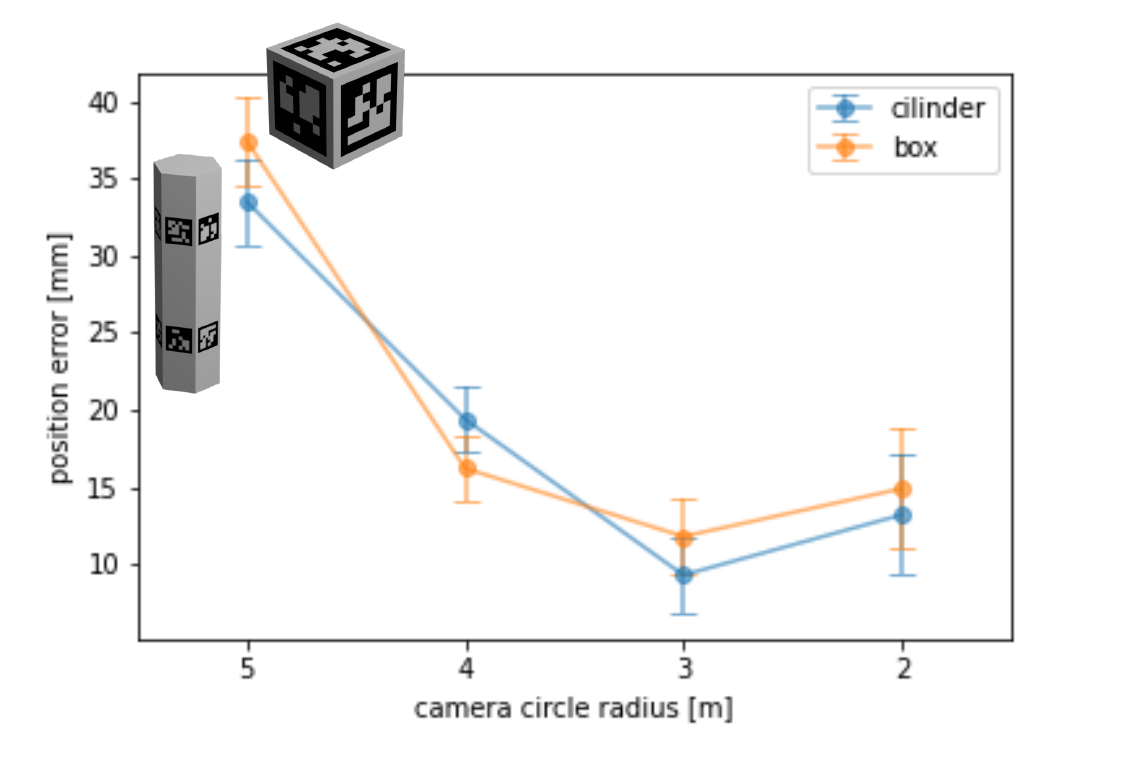}
    \caption{Average position errors in the estimation of the target objects, using four cameras at different distances using two configuration props.}
    \label{Fig:calibration_result}
\end{figure}

As the system's precision is influenced by the distance between the camera and the object (i.e., the farther the object, the lower the accuracy), we repeat the experiment at four different distances using two different shaped patterns designed in Blender. The cameras form circles of radius 2, 3, 4, and 5 meters. The average pose error of the calibrated cameras w.r.t. the original pose spawned is then averaged and plotted in Fig.~\ref{Fig:calibration_result}.

\begin{figure}
    \centering
    \includegraphics[width=\columnwidth]{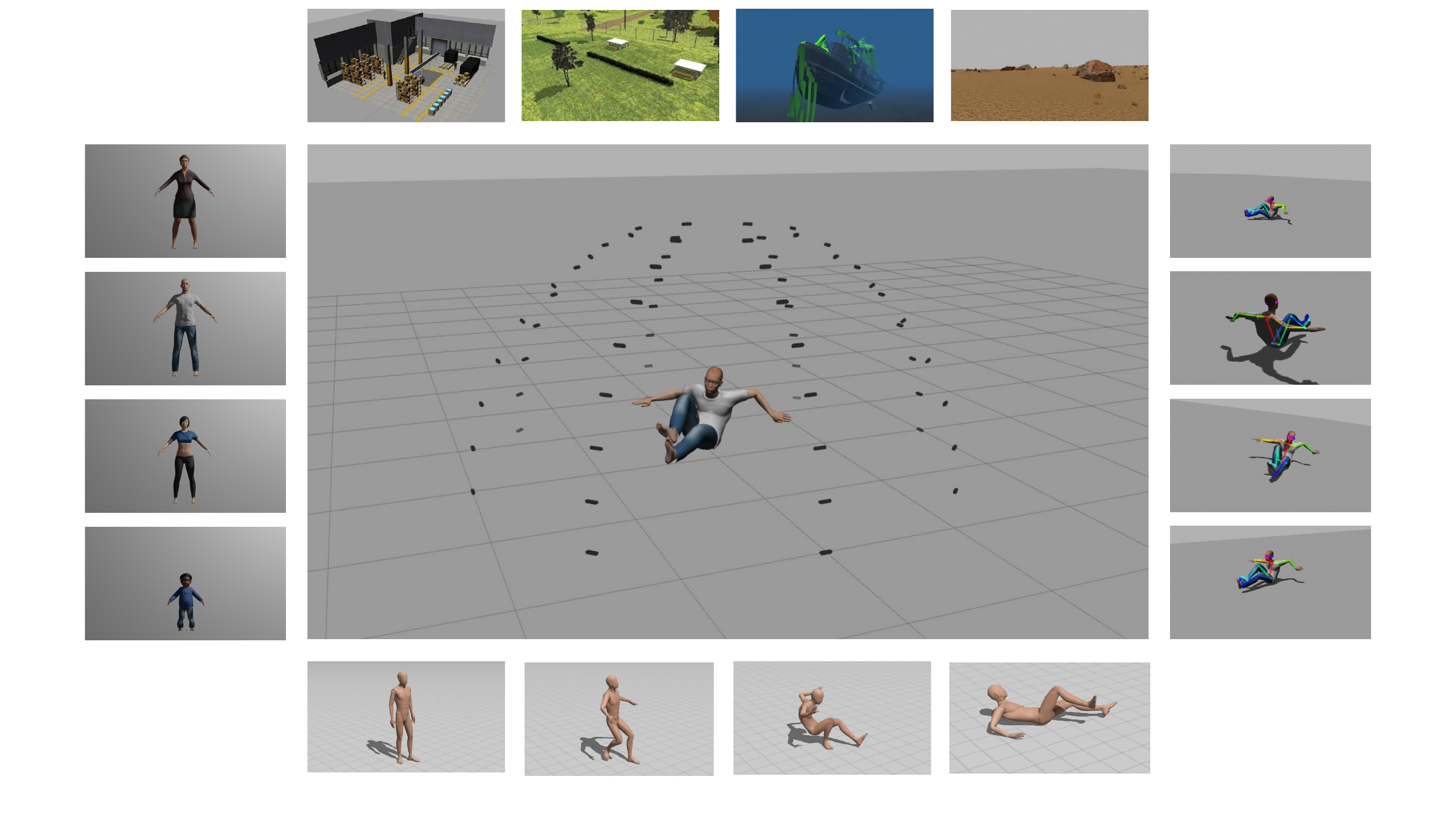}
    \caption{Illustration of the Open-VICO use-case to create synthetic datasets of human motions. The toolkit allows the definition of numerous environments (top), human kinematic models and appearance configurations (left), and human actions and motions (bottom), within a framework of multi-vision systems (center) and human tracking algorithms (right).}
    \label{Fig:dataset}
\end{figure}

\subsection{Synthetic Dataset Creation}
\label{subsec:dataset}

Regarding the creation of synthetic datasets with multi-vision systems, this section considers the scenario described by~\cite{Tsai2019} as an example. In which the dataset ``NTU RGB+D 120''~\cite{Liu2020} for 3D action recognition is used to train a deep learning algorithm for fall detection. Open-VICO's pipeline relies on the MakeHuman ``Mass produce'' plugin to generate large sets of humans and clothes. Each model is retargeted with synthetic falling animations downloaded from the Mixamo~\footnote{\url{https://www.mixamo.com}} website. The models and the animations are then performed within a parametrized vision setup spawned in the gazebo world (see Fig.~\ref{Fig:dataset}). As a result is possible to obtain a huge dataset with countless point of view with little effort.

\subsection{Multi-Camera Human Tracking}
\label{subsec:human_tracking}

\begin{figure}
    \centering
    \includegraphics[trim=7.0cm 1.0cm 7.0cm 1.0cm, clip, width=\columnwidth]{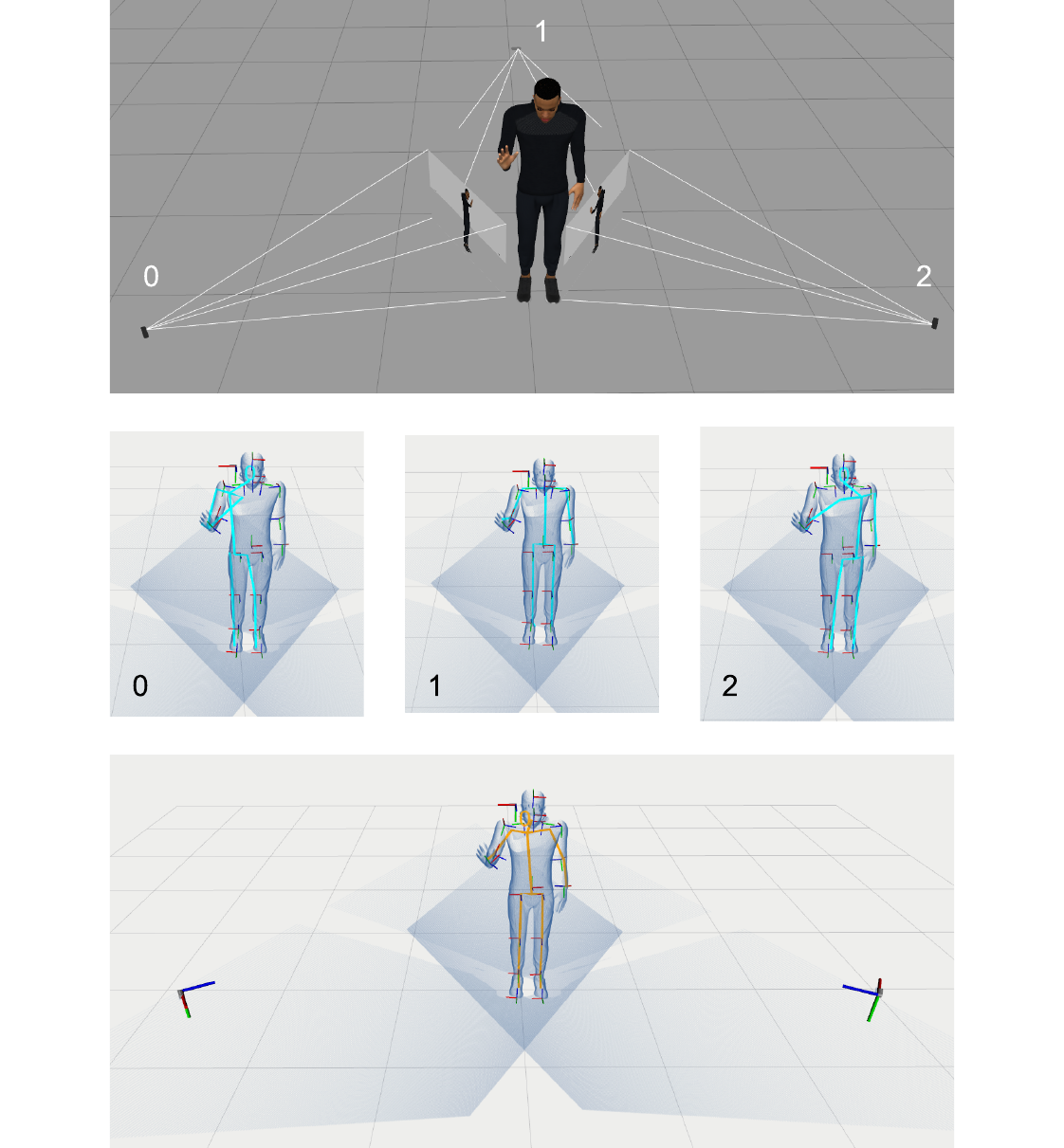}
    \caption{Visualization of multi-camera human tracking in simulation using Open-VICO and OpenPose. Gazebo visualization of the setup (top). Single-camera tracking compared with the ground truth (middle). Fused data tracking compared with the ground truth (bottom).}
    \label{Fig:ground}
\end{figure}

As proof of concept, this section shows the integration of a basic fusion algorithm in Gazebo, benefiting Open-VICO features. An enhanced 3D RGB-D-based OpenPose is implemented exploiting multiple cameras inside the simulator. This example employs 3 RGB-D cameras spawned around a virtual subject on a 3m radius circumference. Fig.~\ref{Fig:ground} shows the overlapping of the depth cloud, the ground truth, and the OpenPose outcomes for a particular camera. 

\begin{figure}
    \centering
    \includegraphics[width=1\columnwidth]{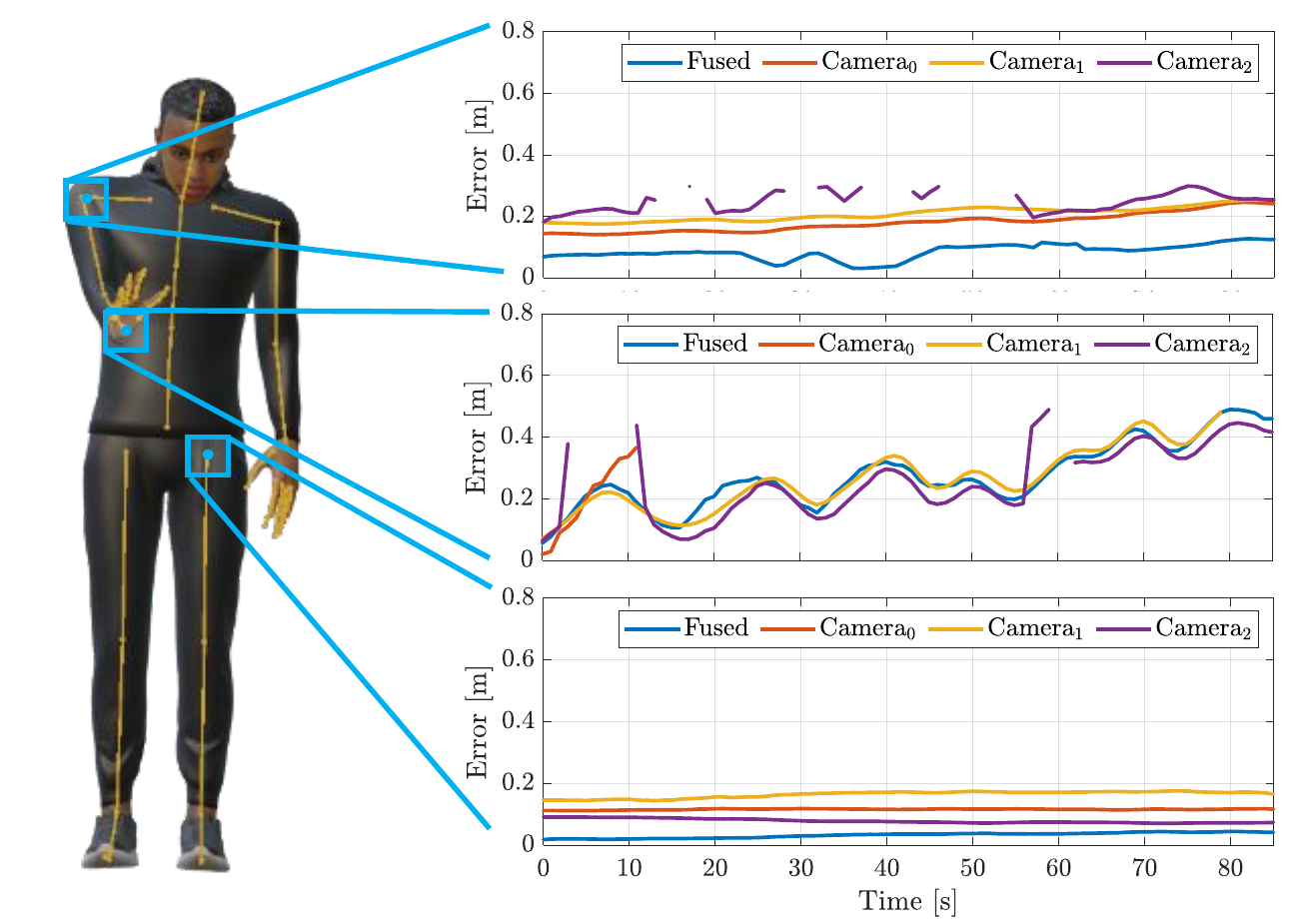}
    \caption{Results of the multi-camera human tracking application in simulation thanks to Open-VICO features. When occlusions occurs, single-cameras cannot view certain joints. However, fused data is robust to occlusions and results are more accurate in every case.}
    \label{Fig:results_nulti}
\end{figure}

\begin{figure*}
    \centering
    \includegraphics[width=0.86\textwidth]{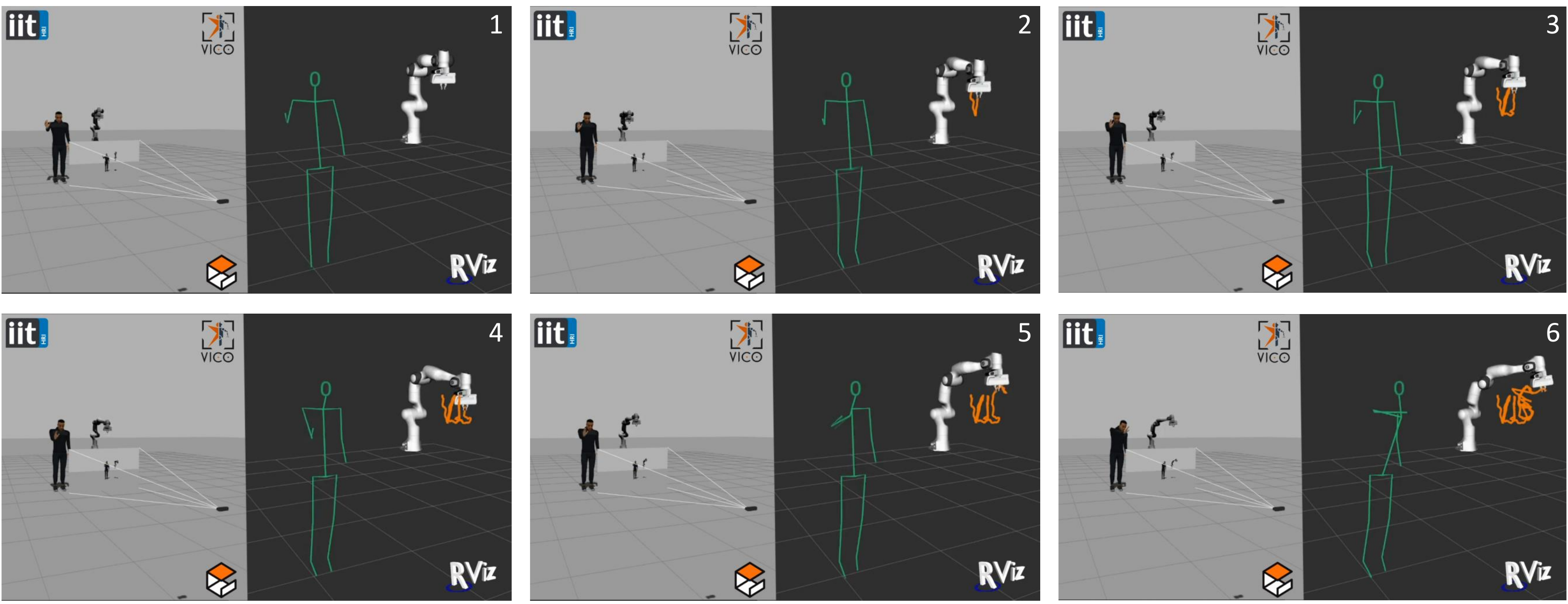}\\
    (a)\\ \vspace{0.3cm}
    \includegraphics[width=0.86\textwidth]{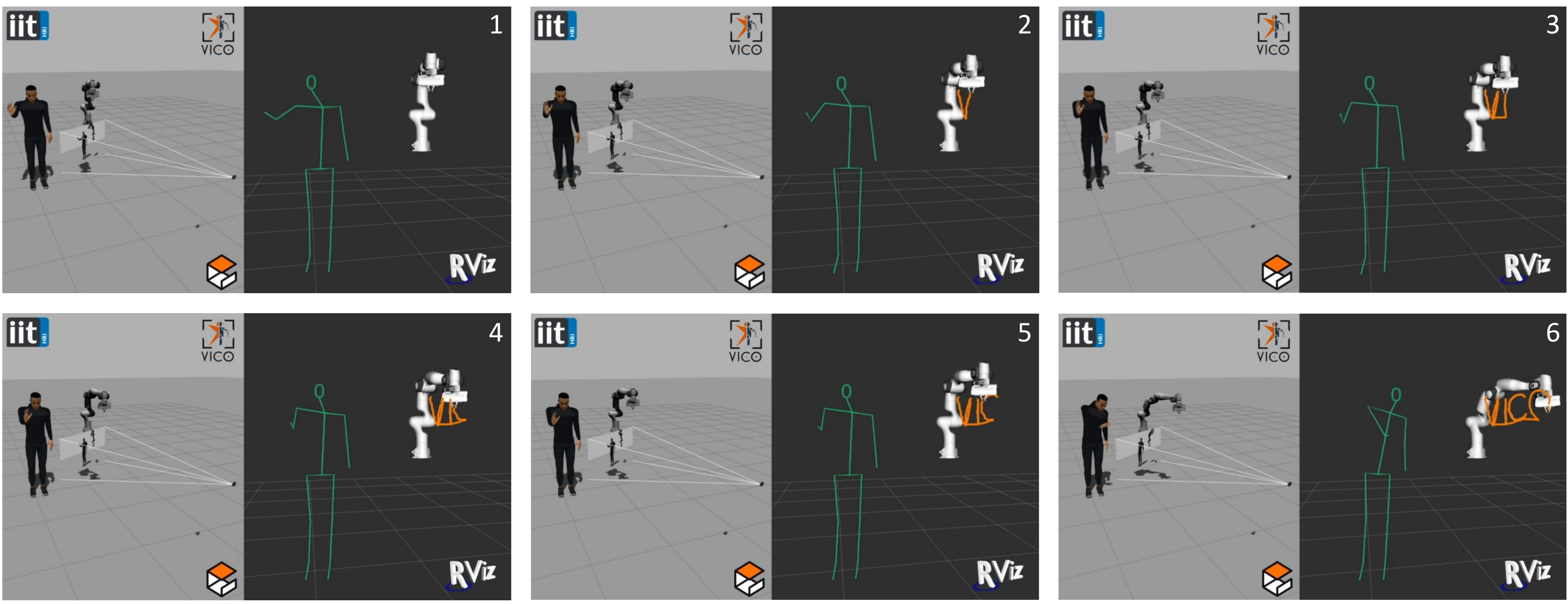}\\
    (b)
    \caption{Screenshots of a demonstration in which a virtual human teleoperates a robotic manipulator attempting to write the world "\textit{VICO}" using OpenPose inside Gazebo thanks to Open-VICO (left: Simulator -- Gazebo visualization, right: RViz visualization). The top sequence depicts a failure case as the camera is far from the virtual human, while the bottom sequence shows a successful case with a camera placed in a better location.}
    \label{Fig:teleop}
	\vspace{-0.5cm}
\end{figure*}

The joints' positions are then logged in a MATLAB-compatible format for further analysis and evaluation. Open-VICO provides the tools to implement and analyze multi-vision systems and algorithms for marker-less human tracking in simulation. In the particular toy example presented in this paper, the fusion of the data perceived from the three OpenPose systems is realised by applying a simple average filter. The results are depicted in Fig.~\ref{Fig:results_nulti}. The analysis allows evaluating the performance of the applied methodologies, detecting errors, and applying enhanced fusion algorithms to improve robustness.



\subsection{Human-Robot Interaction}
\label{subsec:research_examples}

This section presents an example of fully-virtual teleoperation inside Gazebo thanks to the Open-VICO features. The example demonstrates the need to have these systems in simulation to test the camera settings and algorithms while ensuring safety both for the human and the robot.
In the application, a robotic arm reproduces the motion of a human during a writing task attempting to write the world "\textit{VICO}". The motion of the human has been previously recorded in a real scenario with a MoCap system and integrated inside Gazebo according to the Open-VICO process described in~\ref{subsec:human_model}. A Franka Emika manipulator is controlled using a cartesian impedance controller~\cite{Schaffer2003} and the inputs given by 3D Openpose with an RGB-D camera. The software acquires the initial position of the right hand and the end-effector frame w.r.t. the world frame. The following acquisitions are needed to compute the hand displacement from the initial position and send it to the robot controller as desired equilibrium pose.

Two particular cases are presented, in which the only difference is the camera's location in the workspace. In the first case, the system fails, while the second is successful, demonstrating the fragility of markerless vision systems and the potential of Open-VICO for developing safe HRI applications. Fig.~\ref{Fig:teleop} shows how a wrong framing of a camera inputting the 3D OpenPose tracking system may affect the outcome writing task.

%% file: ms5.tex
In conclusion, this paper proposes a comprehensive collection and integration of tools in the Open-VICO toolkit as the first of its kind to the authors' best knowledge. The framework allows the integration of human-simulated models into the Gazebo robotic simulator environment and provides the potential to test and verify HRI systems. Flexibility in package parameters like the simulated world, recorded motion plans, and camera parameters allowed for rapid generation of scenarios. The software architecture and potential applications were described along with the whole toolkit pipeline to create virtual human kinematic models and motions and integrate them in a Gazebo world. Moreover, four use-cases were presented, demonstrating the toolkit's potential for multi-sensor vision systems and HRI in simulation.
Future works will consider including collision features to enhance the collaboration tasks between actors. An online control marionette-like of the human model in Gazebo using a motion capture system will be proposed. Furthermore, we will enlarge and integrate the panorama of deep-learning-based mocap systems to offer a more extensive selection for richer cross-comparison.

%% file: ms6.tex
This work was supported in part by the European Union’s Horizon 2020 research and innovation program under Grant Agreement No. 871237 (SOPHIA) in part by the ERC-StG Ergo-Lean (Grant Agreement No.850932).